\documentclass[10pt,twocolumn,letterpaper]{article}

\usepackage{cvpr}
\usepackage{times}
\usepackage{epsfig}
\usepackage{graphicx}
\usepackage{amsmath}
\usepackage{amssymb}
\usepackage{caption}
\usepackage{enumitem}   

\graphicspath{ {images/} }

\usepackage[breaklinks=true,bookmarks=false]{hyperref}

\cvprfinalcopy 

\def\cvprPaperID{****} 

\title{3D Reconstruction from Sketches}

\author{Abhimanyu Talwar\\
Harvard University\\
{\tt\small abhimanyutalwar@g.harvard.edu}
\and
Julien Laasri\\
Harvard University\\
{\tt\small jlaasri@g.harvard.edu}
}

\begin{document}
\twocolumn[{%
\renewcommand\twocolumn[1][]{#1}%
\maketitle
\vspace{-2.5em}
\begin{center}
    {\small December 12, 2018}
\end{center}
\begin{center}
    \centering
    \includegraphics[width=15cm]{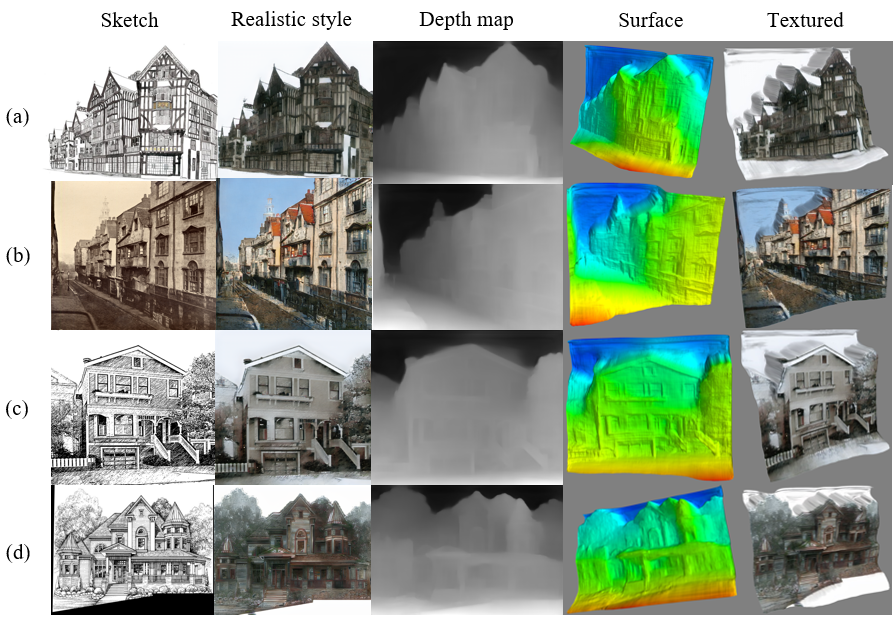}
    \captionof{figure}{All steps of our pipeline for different drawings. Only sketch (d) required the stitching process.}
    \label{fig:main}
\end{center}%
}]

\maketitle

\begin{abstract}
   We consider the problem of reconstructing a 3D scene from multiple sketches. We propose a pipeline which involves (1) stitching together multiple sketches through use of correspondence points, (2) converting the stitched sketch into a realistic image using a CycleGAN, and (3) estimating that image's depth-map using a pre-trained convolutional neural network based architecture called MegaDepth. Our contribution includes constructing a dataset of image-sketch pairs, the images for which are from the Zurich Building Database, and sketches have been generated by us. We use this dataset to train a CycleGAN for our pipeline's second step. We end up with a stitching process that does not generalize well to real drawings, but the rest of the pipeline that creates a 3D reconstruction from a single sketch performs quite well on a wide variety of drawings.
\end{abstract}

\section{Introduction}

\begin{figure}[!h]
\centering
\includegraphics[width=8.0cm]{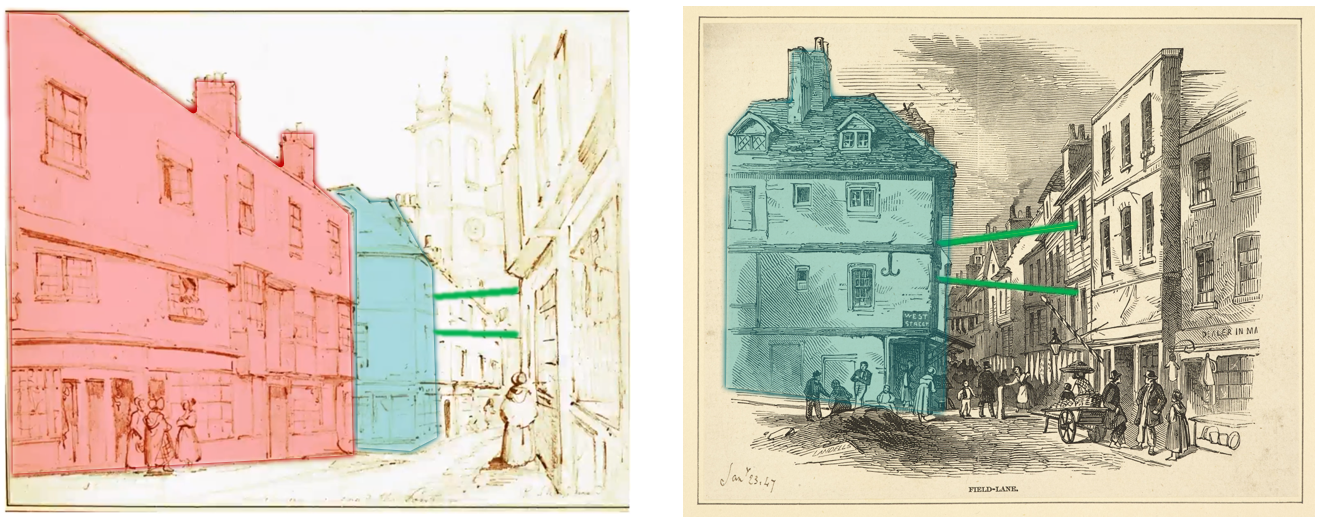}
\caption{Two neighbouring viewpoints of Field Lane at close time periods. The building highlighted in red was demolished between the two drawings. Note the same building highlighted in blue and the timbers highlighted in green that appear in the two drawings. Finding a way to stitch parts of these drawings together would allow us to have the refined details of the drawing on the right with the inclusion of the building that was destroyed from the drawing on the left.}
\label{justification_stitching}
\end{figure}

Reconstructing scenes in 3D from multiple photographic images is an active area of research in Computer Vision. This could potentially allow us to recreate 3D models of streets or entire cities based on image or video inputs. Our goal was to recreate a 3D model of an early 1800s Field Lane, a notorious street of London described by Charles Dickens in Oliver Twist. However, we do not have images of this scene as cameras did not exist back then. The only graphical resources that we have are a few drawings from different artists. In this paper, we propose a pipeline to obtain a 3D reconstruction of a street drawn from multiple viewpoints in different sketches. Such a pipeline could allow us to reconstruct 3D scenes of the pre-photograph era and recreate imaginary settings drawn by different artists. Three main problems are tackled in this paper: how to find correspondences and combine multiple sketches that describe the same scene from different viewpoints (see example use case on Field Lane in figure \ref{justification_stitching}), how to obtain a realistic output from simple sketches, and how to infer depth from 2D images.

\section{Related work}

Finding correspondences in images is a common Computer Vision problem.The main issue is the identification of important regions of each image and the creation of significant features from them. Many features have been developed over the years, among the most popular ones we find SIFT \cite{sift}, SURF \cite{surf} and ORB \cite{orb}.

Getting realistic looking images from drawings is a style transfer process. A lot of research has been focused on these techniques during the last years. Popular methods that can be applied to our problem include Image-to-Image translations \cite{isola2017image} and the usage of CycleGANs \cite{CycleGAN2017}.

Reconstruction of a 3D view from one or more sketches or images is an old problem, and several researchers have proposed effective solutions in the past. In \cite{liebowitz99}, Liebowitz, Criminisi and Zisserman utilised geometric features present in Renaissance paintings of architecture (such as parallel and perpendicular lines) to recreate them in 3D. Saxena, Sun and Ng have proposed a supervised learning method in Make3D \cite{make3d}, in which they train a Markov Random Field to estimate a depth-map from a single image.

Inferring depth from a single image is currently one of the hottest topics in Computer Vision. As, these solutions are often developed to improve automatic vehicles, they often give the best results in urban settings. This is exactly what we are focusing on in this paper. Among those which give the best results, we got particularly interested in two Deep Learning techniques: Monodepth \cite{monodepth} and MegaDepth \cite{megadepth}.

\section{Approach}
\begin{figure}[!h]
\centering
\includegraphics[width=8.0cm]{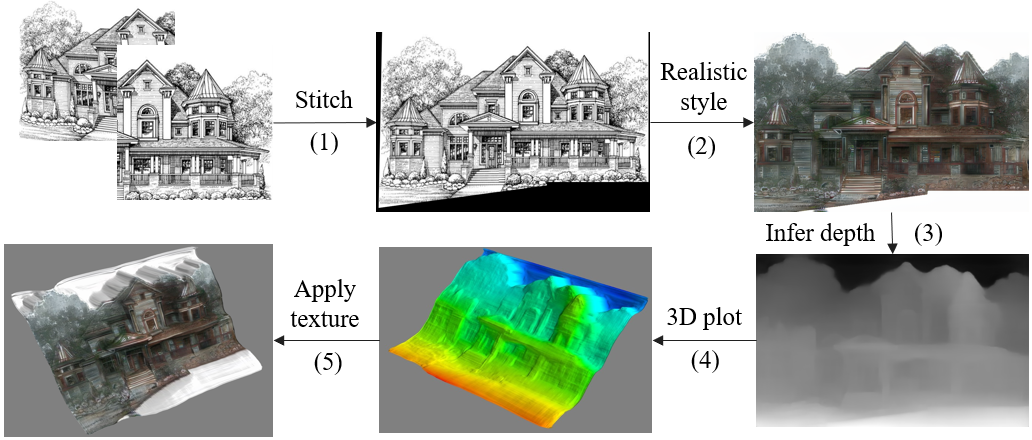}
\caption{Our entire pipeline takes in multiple sketches of the same scene with correspondences and outputs a 3D reconstruction of the scene.}
\label{pipeline}
\end{figure}

As shown in figure \ref{pipeline}, the entire pipeline we came up with takes as input multiple sketches of the same scene from different viewpoints and outputs a 3D reconstruction of the scene. If we only want the 3D reconstruction of a single drawing, we may skip the stitching part (1) of the pipeline.

Our entire pipeline comprises 5 steps:
\begin{enumerate}[label=(\arabic*)]
    \item Sketches which have correspondences can be stitched together. We can keep on stitching new sketches to the stitched drawing as long as they are correspondences between the images. We can then treat each stitched image independently for the rest of the pipeline.
    \item We use a CycleGAN\cite{CycleGAN2017} to convert sketches to realistic-looking images.
    \item We infer depth from the realistic looking image using MegaDepth, a pre-trained Neural Network \cite{megadepth}.
    \item We do a surface plot in 3D.
    \item We use the realistic-looking image as a texture for the surface plot.
\end{enumerate}

Steps (3), (4) and (5) ended up being quite straightforward. We used pre-trained MegaDepth as is to infer depth from a realistic-looking image. We then used the Python package Mayavi \cite{mayavi} to complete steps (4) and (5) simultaneously and quite easily. Thus, we focused our efforts on steps (1) stitching and (2) realistic style transfer of our pipeline.

\subsection{Stitching drawings together}

\begin{figure}[!h]
\centering
\includegraphics[width=8.0cm]{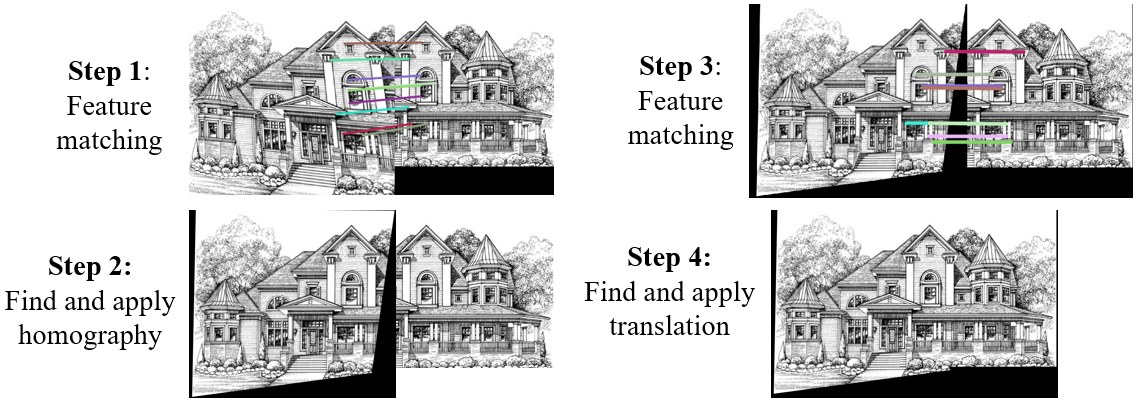}
\caption{Process to stitch multiple drawings of the same scene together.}
\label{stitching}
\end{figure}

The process we introduce to stitch two drawings together comprises 4 steps, as shown in figure \ref{stitching}. We assume that the two drawings have correspondences. For instance, the same building can appear in two different drawings even though the viewpoints of the scene may differ. This is the situation that we encounter when dealing with Field Lane drawings. The process is as follows.
\begin{enumerate}
    \item We find and match features of the two drawings.
    \item We find the homography that transforms the features of drawing 1 to their matched features in drawing 2. We then apply this homography to the whole drawing 1. In other words, if these drawings were actual photos taken with perspective cameras, this corresponds to getting the view of drawing 1 from the camera position and angle of drawing 2 as describe by Szeliski in Section 2.1.5, Mapping from one camera to another, of his Computer Vision book \cite{cvbook}.
    \item We find and match the features of this new perspective with the features of drawing 2.
    \item We compute the mean of the translations that map pairs of matched features together when the two drawings are put side by side. We apply that translation to put drawing 2 on top of drawing 1.
\end{enumerate}

A crucial element of this process is the definition of features for these drawings. These should be defined from the identification of local regions that are significant in an image. In addition to that, we need these features to be invariant by change of scale and perspective. This would allow our process to identify the same ones in two different viewpoints of the same scene. The commonly used approaches to face this issue include SIFT\cite{sift} and SURF\cite{surf} features. However, as these methods are patented, we have decided to opt for ORB\cite{orb} features as this unpatented technique also aims at dealing with these constraints and gives pretty decent results on our toy data as can be seen in figure \ref{stitching}.

\subsection{From Sketch to Image using CycleGANs}
We used a CycleGAN to translate sketches into images. This is an important step in our pipeline because (1) it makes our 3D renderings look realistic, and (2) it allows us to use pre-trained models trained on real images (such as MegaDepth) for inferring depth. A CycleGAN requires two sets for training, one containing sketches and the other with images. These sets need not contain paired images. Our key contribution for this report has been creation of a set of sketches from images contained in the Zurich Building Image Database \cite{10.1007/3-540-45113-7_8}. We then used those sketches along with the images to train a CycleGAN. 

\begin{figure}[!h]
\centering
\includegraphics[width=8.0cm]{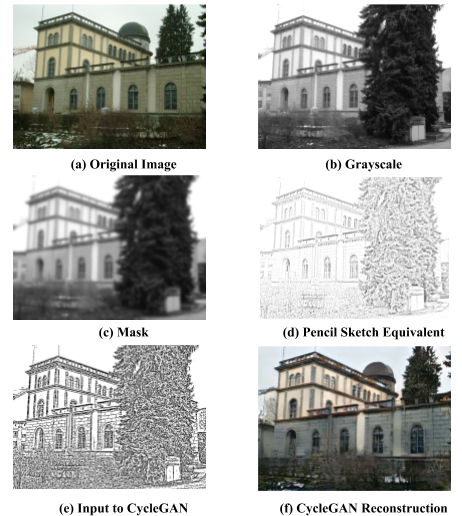}
\caption{Steps involved in creation of our dataset of sketches of buildings from images in the Zurich Building Image Database.}
\label{sketches}
\end{figure}

The steps involved in creating sketches from images are as follows. We have used an image processing technique called "Dodging" (as described by Michael Beyeler in \cite{Pencil}), to create pencil sketch equivalents of images (as shown in Fig. \ref{sketches}(d)).

\begin{enumerate}
    \item We begin with image of a building (such as Fig. \ref{sketches}(a)), and convert it to grayscale (Fig. \ref{sketches}(b)).
    \item To implement Dodging, we blur the inverted grayscale (Fig. \ref{sketches}(c)) and invert it back to get a mask (Fig. \ref{sketches}(c)). We then divide (pixel-wise), the intensities of the original grayscale by the mask's, and multiply by a scale-factor of $256$, to get a pencil sketch look-alike (Fig. \ref{sketches}(d)).
    \item We observed that most of the sketches available online were noisier than our pencil sketch. To correct our pencil sketch, we put it through a high-pass filter, and took its negative to get Fig. \ref{sketches}(e), which closely resembled the form of sketches available online.
    \item We trained our CycleGAN using a set of images and sketches generated using the above methodology (images as shown in Fig. \ref{sketches}(a) and sketches as shown in Fig. \ref{sketches}(e)). Once trained, our CycleGAN's generator model reconstructed Fig. \ref{sketches}(f) from Fig. \ref{sketches}(e).
\end{enumerate}

\subsection{Experiments with CycleGAN}
\subsubsection{Training Setup}
As detailed in Section 3.2 above, we created a set of sketches from the $1,005$ images of buildings contained in the Zurich Building Image Database \cite{10.1007/3-540-45113-7_8}. Since this is a paired dataset (each image has its corresponding sketch), to reduce the chances of overfitting, we randomly chose a subset of $600$ images, and independently chose $600$ sketches to create our training set. Our hyperparameter settings are detailed below:
\begin{enumerate}
    \item Architecture: We used the off-the-shelf CycleGAN architecture made available by the authors of the original paper \cite{CycleGAN2017}.
    \item Regularization Scheme: We used random crops for data augmentation. We did not use other techniques (such as color jitter, or rotation) because we wanted the Generator to learn realistic looking images (in which buildings stand upright, and colors are natural). In the default CycleGAN settings, images are resized to $286 \times 286$ and then a random crop of size $256 \times 256$ is taken, which means we discard away 20\% of the pixels. To increase the amount of regularization (because our training set of $600$ images per class is smaller than the usual), we resized the images to $400 \times 400$ and took random crops of size $320 \times 320$, thereby randomly discarding nearly 36\% pixels for each training example. 
    \item Number of Epochs: We trained the CycleGAN (starting with randomly initialized weights) for $200$ epochs.
    \item Learning Rate Regime: We used the default learning rate regime which trains at $lr=0.0002$ for the first $100$ epochs and then linearly decrease the rate to $0$ over the next $100$ epochs.
\end{enumerate}

\subsubsection{Test Time}
We describe below the settings we used for reconstructing realistic images from sketches using our trained CycleGAN generator.
\begin{enumerate}
    \item Pre-processing: Fig. \ref{fig:main} (under the header "Sketches") shows some of the sketches from our Test Set. We did not feed them directly to our Generator - instead we first processed them by running our sketch generation method detailed in Section 3.2 above, and then fed the processed images to the Generator.
    \item Output Size: Even though we had trained our CycleGAN on crops of size $320 \times 320$, during Testing, we specified the output size to be $480 \times 480$ (by tweaking the --fineSize setting). We observed that we were able to get decent results for this larger size. Fig. \ref{fig:main} (under the header "Realistic Style") shows the reconstructions produced on test set sketches by our trained Generator.
\end{enumerate}

\section{Experimental Results}

\subsection{The stitching procedure}
\begin{figure}[!h]
\centering
\includegraphics[width=8.0cm]{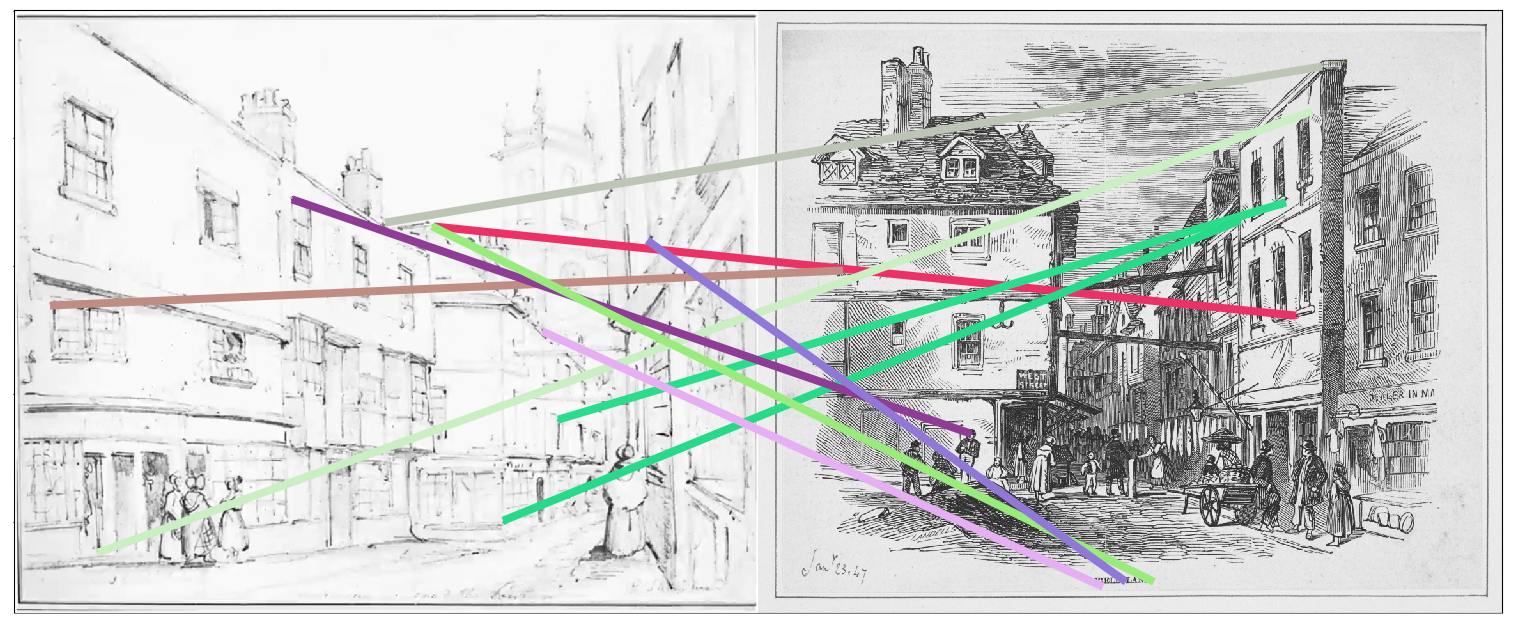}
\caption{Feature matching completely fails on the two perspectives of Field Lane from figure \ref{justification_stitching}.}
\label{matching_fails}
\end{figure}

The stitching procedure can work well on really precise sketches that use the same style and are really just a homography apart from each other. That is why it works on our toy data of figure \ref{stitching} that was built by separating an existing sketch into two overlapping sketches and applying a random homography to one of the resulting drawings. However, on real data such as the one that we have for Field Lane of figure \ref{justification_stitching}, feature matching completely fails as shown in figure \ref{matching_fails}.

\begin{figure}[!h]
\centering
\includegraphics[width=8.0cm]{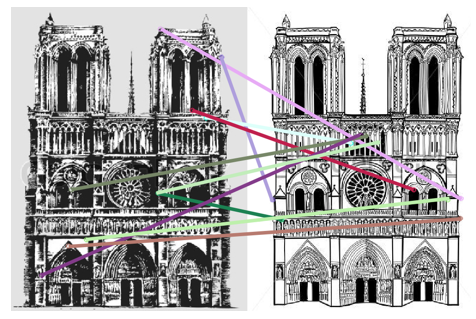}
\caption{Feature matching fails for two sketches of the same building under the same perspective but with different drawing styles.}
\label{cath}
\end{figure}

We may acknowledge that the feature matching of these two particular drawings is quite difficult and even a human would have trouble finding correspondences without the highlights of figure \ref{justification_stitching}, especially due to the disappearance of the red building from one drawing to the other. However, we tried feature matching on two drawings that only differ in style, not much in perspective. This task would be quite easy for a human, but it is not completed well with our feature matching procedure as shown in figure \ref{cath}. The styles of the drawings that we have are too different for our stitching to work well.

On the contrary of images, even for the same scene seen from the same perspective, local areas of drawings from different artists differ a lot. Details emphasized by artists differ from one artist to another and are thus not consistent from one drawing to another. That is why our stitching strategy only works well if the drawings are extremely precise and the styles are exactly the same in the two drawings, which is the case of our toy data from figure \ref{stitching}. Still, we can run the rest of the pipeline on individual sketches, which is what we do in Section \ref{rest}.
 
\subsection{Rest of the pipeline}\label{rest}

\begin{figure}[!h]
\centering
\includegraphics[width=8.0cm]{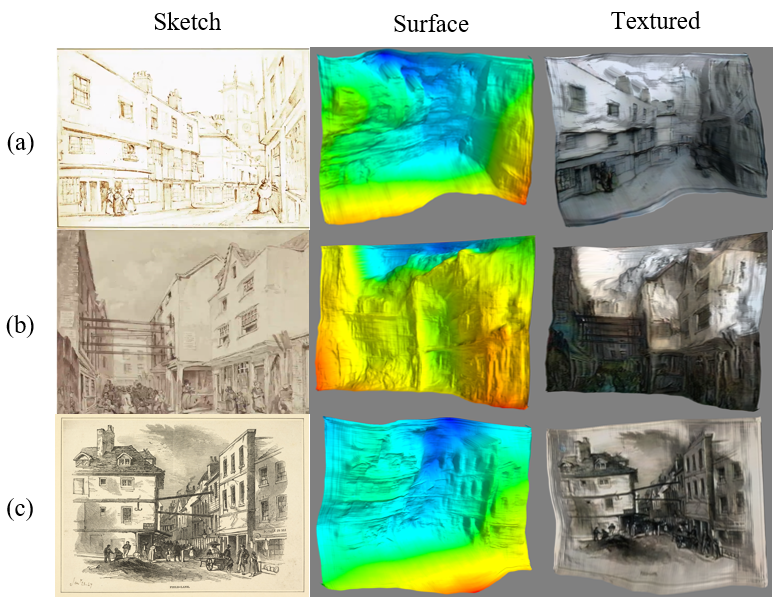}
\caption{Rest of the pipeline applied to sketches of Field Lane.}
\label{results_fieldlane}
\end{figure}

Running the rest of the pipeline on different sketches of Field Lane gives us the results of figure \ref{results_fieldlane}.
 
The CycleGAN doesn't always give such a realistic look to our drawings. For instance, drawing (c) of figure \ref{results_fieldlane} still looks like a drawing after reconstruction, and drawing (a) did not get much color from the style transfer. Still, it seems that the 3D reconstruction that we get for these sketches are quite descent.

We then tried our pipeline on different drawings that have nothing to do with Field Lane. We get much better results with some of them as shown in figure \ref{fig:main}. Our CycleGAN accurately distinguishes the roof from the facade and colors them properly for drawings (a) and (b). Even when the colorization is not that vivid, which is the case of drawing (c), the reconstruction that we get is quite realistic. We also successfully ran our entire pipeline on our toy data (d), including the stitching process, and it gives pretty descent results. Overall, the 3D reconstruction seems to capture quite a big amount of details from the realistic-looking image, and we can specifically see that on the depth maps of figure \ref{fig:main}.
 
\section{Conclusion}

In this paper, we addressed the problem of reconstructing a 3D scene from multiple sketches drawn from different angles. We proposed an end-to-end pipeline to accomplish this. Even though the stitching process does not appear to work well on real imprecise sketches with different styles, we ended up with a solid single-sketch-to-image pipeline that gives pretty descent results on a wide variety of drawings as shown in figures \ref{fig:main} and \ref{results_fieldlane}.

We have a couple of ideas that could help us improve the results that we get on Field Lane and on other different drawings.

\begin{itemize}
    \item With the corresponding resources needed to accomplish that, we could ask artists with different styles to draw particular photos of some buildings. Our CycleGAN could thus try to map their drawings to the realistic photos of the buildings we asked them to draw. Having different artists with different styles would probably make our CycleGAN more robust to changes of style.
    \item A more realistic way of making our CycleGAN more robust to changes of style is varying the process that we use to create sketches from real images. Different processes would lead to different drawing styles, thus augmenting our distribution of drawings of our CycleGAN.
    \item Another idea would be to fine tune MegaDepth to the output of our CycleGAN using stereo images. This may allow us to get even better 3D reconstructions from our realistic-looking images as MegaDepth would then specialize to our CycleGAN's outputs.
\end{itemize}
 
{\small
\bibliographystyle{unsrt}
\bibliography{egbib}

\begin{thebibliography}{10}

\bibitem{sift}
David~G. Lowe.
\newblock Distinctive image features from scale-invariant keypoints.
\newblock {\em International Journal of Computer Vision}, 60(2):91--110, Nov
  2004.

\bibitem{surf}
Herbert Bay, Tinne Tuytelaars, and Luc Van~Gool.
\newblock Surf: Speeded up robust features.
\newblock In Ale{\v{s}} Leonardis, Horst Bischof, and Axel Pinz, editors, {\em
  Computer Vision -- ECCV 2006}, pages 404--417, Berlin, Heidelberg, 2006.
  Springer Berlin Heidelberg.

\bibitem{orb}
E.~Rublee, V.~Rabaud, K.~Konolige, and G.~Bradski.
\newblock Orb: An efficient alternative to sift or surf.
\newblock In {\em 2011 International Conference on Computer Vision}, pages
  2564--2571, Nov 2011.

\bibitem{isola2017image}
Phillip Isola, Jun-Yan Zhu, Tinghui Zhou, and Alexei~A Efros.
\newblock Image-to-image translation with conditional adversarial networks.
\newblock In {\em Computer Vision and Pattern Recognition (CVPR), 2017 IEEE
  Conference on}, 2017.

\bibitem{CycleGAN2017}
Jun-Yan Zhu, Taesung Park, Phillip Isola, and Alexei~A Efros.
\newblock Unpaired image-to-image translation using cycle-consistent
  adversarial networks.
\newblock In {\em Computer Vision (ICCV), 2017 IEEE International Conference
  on}, 2017.

\bibitem{liebowitz99}
D.~Liebowitz, A.~Criminisi, and A.~Zisserman.
\newblock Creating architectural models from images.
\newblock In {\em Annual Conference of the European Association for Computer
  Graphics (Eurographics)}, volume~18, pages 39--50, 1999.

\bibitem{make3d}
Ashutosh Saxena, Min Sun, and Andrew~Y. Ng.
\newblock Make3d: Learning 3d scene structure from a single still image.
\newblock {\em IEEE Trans. Pattern Anal. Mach. Intell.}, 31(5):824--840, May
  2009.

\bibitem{monodepth}
Cl{\'{e}}ment Godard, Oisin {Mac Aodha}, and Gabriel~J. Brostow.
\newblock Unsupervised monocular depth estimation with left-right consistency.
\newblock {\em CoRR}, abs/1609.03677, 2016.

\bibitem{megadepth}
Zhengqi Li and Noah Snavely.
\newblock Megadepth: Learning single-view depth prediction from internet
  photos.
\newblock {\em CoRR}, abs/1804.00607, 2018.

\bibitem{mayavi}
P.~Ramachandran and G.~Varoquaux.
\newblock {Mayavi: 3D Visualization of Scientific Data}.
\newblock {\em Computing in Science \& Engineering}, 13(2):40--51, 2011.

\bibitem{cvbook}
Richard Szeliski.
\newblock {\em Computer Vision: Algorithms and Applications}.
\newblock Springer-Verlag, Berlin, Heidelberg, 1st edition, 2010.

\bibitem{10.1007/3-540-45113-7_8}
Hao Shao, Tom{\'a}{\v{s}} Svoboda1, Tinne Tuytelaars, and Luc Van~Gool.
\newblock Hpat indexing for fast object/scene recognition based on local
  appearance.
\newblock In Erwin~M. Bakker, Michael~S. Lew, Thomas~S. Huang, Nicu Sebe, and
  Xiang~Sean Zhou, editors, {\em Image and Video Retrieval}, pages 71--80,
  Berlin, Heidelberg, 2003. Springer Berlin Heidelberg.

\bibitem{Pencil}
Michael Beyeler.
\newblock How to create a beautiful pencil sketch effect with opencv and
  python.
\newblock
  \url{http://www.askaswiss.com/2016/01/how-to-create-pencil-sketch-opencv-python.html}.
\newblock Accessed: 2018-12.

\end{thebibliography}
}

\end{document}